\def\year{2018}\relax
\documentclass[letterpaper]{article} %DO NOT CHANGE THIS
\usepackage{aaai18}  %Required
\usepackage{times}  %Required
\usepackage{helvet}  %Required
\usepackage{courier}  %Required
\usepackage{url}  %Required
\usepackage{graphicx}  %Required
\usepackage{latexsym}
\usepackage{CJKutf8}
\usepackage{booktabs}
\usepackage{amsmath}
\usepackage{amssymb}
\usepackage{enumitem}
\usepackage{avisil}
\usepackage{graphicx}
\usepackage{multirow}
\usepackage{xcolor}
\usepackage{tablefootnote}
\usepackage{stmaryrd}
\usepackage{verbatim}
\usepackage{caption}
\usepackage{subfig}

\DeclareMathOperator{\cosine}{cos}

\definecolor{cadmiumgreen}{rgb}{0.0, 0.50, 0.31}
\definecolor{darkgreen}{rgb}{0.0, 0.2, 0.13}
\definecolor{mahogany}{rgb}{0.75, 0.25, 0.0}
\newcommand{\ct}{S}
\newcommand{\wpf}{T}
\newcommand{\cti}{s}
\newcommand{\wpfi}{t}

\begin{document}
% The file aaai.sty is the style file for AAAI Press 
% proceedings, working notes, and technical reports.
%
\title{Neural Cross-Lingual Entity Linking}
\author{Avirup Sil \and Gourab Kundu \and Radu Florian \and Wael Hamza\\
IBM Research AI\\
1101 Kitchawan Road\\
Yorktown Heights, NY 10598\\
\{avi, gkundu, raduf, whamza\}@us.ibm.com
}
\maketitle
\begin{abstract}
A major challenge in Entity Linking (EL) is making effective use of contextual information to disambiguate mentions to Wikipedia that might refer to different entities in different contexts. The problem exacerbates with cross-lingual EL which involves linking mentions written in non-English documents to entries in the English Wikipedia: to compare textual clues across languages we need to compute similarity between textual fragments across languages. In this paper, we propose a neural EL model that trains fine-grained similarities and dissimilarities between the query and candidate document from multiple perspectives, combined with convolution and tensor networks. Further, we show that this English-trained system can be applied, in zero-shot learning, to other languages by making surprisingly effective use of multi-lingual embeddings. The proposed system has strong empirical evidence yielding state-of-the-art results in English as well as cross-lingual: Spanish and Chinese TAC 2015 datasets.

\end{abstract}

\section{Introduction}
% We live in an era of information explosion, being inundated in a vast
% amount of data in various forms: text, video, audio. Being able to automatically analyze it usually involves filling relational tables,
% requiring a processing system to be able
% to uniquely identify actors, agents, places, events, etc. 
Entity Linking (EL) is the task of associating a specific textual mention of an entity (henceforth query entity) in
a given document (henceforth query document) with an entry in a large target catalog of entities, often
called a knowledge base or KB, and is one of the major tasks in the
Knowledge-Base Population (KBP) track at the Text Analysis Conference (TAC) \cite{ji2014overview,ji2015overview}.
Most of the previous EL research \cite{2007-emnlp-conll-cucerzan-wikipedia-NE-disambig,2011-acl-illinois-wikifier,cikm-joint-nerel} have used Wikipedia as the target catalog of entities, because of its coverage and frequent updates made by the community of users. 

Some ambiguous cases for entity linking require computing fine-grained similarity between the context of the query mention and the title page of the disambiguation candidate. Consider the following examples:\\
$e_1$: Alexander Douglas Smith is an American football quarterback for the Kansas City Chiefs of the National Football League (NFL).\\
$e_2$: Edwin Alexander ``Alex" Smith is an American football tight end who was drafted by the Tampa Bay Buccaneers in the third round of the 2005 NFL Draft.\\
$e_3$: Alexander Smith was a Scottish-American professional golfer who played in the late 19th and early 20th century.\\
$q$: Last year, while not one of the NFL’s very best quarterbacks, Alex Smith did lead the team to a strong 12-4 season.

Here, $e_1$, $e_2$ and $e_3$ refer to the Wikipedia pages of three sportsmen (only first sentence is shown), known as ``Alex Smith"; $q$ refers to the sentence for the query mention ``Alex Smith". Since words in $e_3$ belong to a different domain (golf) than $q$ (American football), simple similarity based methods \eg\ TF-IDF based cosine similarity will have no difficulty in discarding $e_3$ as disambiguation for $q$. But words in $e_1$ and $e_2$ contain significant overlap (both are American football players) even in key terms like NFL. Since ``Alex Smith'' in $q$ is a quarterback, correct disambiguation for $q$ is $e_1$. This requires fine-grained similarity computation between $q$ and the title page of $e_1$. In this paper, we propose \textit{training} state-of-the-art (SOTA) similarity models between the context of the query mention and the page of the disambiguation candidate from Wikipedia such that the similarity models can learn to correctly resolve such ambiguous cases. We investigate several ways of representing both the similarity and coherence between the query document and candidate Wikipedia pages. For this purpose, we extract contextual information at different levels of granularity using the entity coreference chain, as well as surrounding mentions in the query document, then use a combination of convolutional neural networks (CNN), LSTMs \cite{hochreiter1997long}, Lexical Composition and Decomposition \cite{wang2016sentence},  Multi-Perspective Context Matching (MPCM) \cite{wang2016multi}, and Neural Tensor Networks  \cite{socher2013reasoning,socher2013tensor} to encode this information and ultimately perform EL.

The TAC community is also interested in cross-lingual EL \cite{tsai2016cross,sil2016one}: given a mention in a foreign language document \eg\ Spanish or Chinese, one has to find its corresponding link in the English Wikipedia. The main motivation of the task is to do Information Extraction (IE) from a foreign language for which we have extremely limited (or possibly even no) linguistic resources and no machine translation technology. The TAC 2017 pilot evaluation \footnote{http://nlp.cs.rpi.edu/kbp/2017/taskspec\_pilot.pdf} targets really low-resource languages like Northern Sotho or Kikuyu which only have about 4000 Wikipedia pages which is a significantly smaller size than the English Wikipedia. Recently, for cross-lingual EL, \cite{tsai2016cross} proposed a cross-lingual wikifier that uses multi-lingual embeddings. However, their model needs to be re-trained for every new language and hence is not entirely suitable/convenient for the TAC task. We propose a zero shot learning technique \cite{palatucci2009zero,socher2013zero} for our neural EL model: once trained in English, it is applied for cross-lingual EL without the need for re-training. We also compare three popular multi-lingual embeddings strategies and perform experiments to show which ones work best for the task of zero-shot cross-lingual EL. The results show that our methods not only obtain results that are better than published SOTA results on English, but it can also be applied on cross-lingual EL on Spanish and Chinese standard datasets, also yielding SOTA results.

\section{Entity Linking Formulation}
We formalize the problem as follows: we are given
a document $D$ in any language, a set of mentions
$M_D = {m_1;\ldots;m_n}$ in $D$, and the English
Wikipedia. For each mention in the document, the
goal is to retrieve the English Wikipedia link that the
mention refers to. If the corresponding entity or concept
does not exist in the English Wikipedia, ``NIL”
should be the answer.

Given a mention $m \in M_D$, the first step is to generate
a set of link candidates $L_m$. The goal of this
step is to use a {\em fast match} procedure to obtain a list of links which hopefully include the correct answer. We only look at the surface
form of the mention in this step, and use no contextual
information.
The second essential step is the ranking step where
we calculate a score for each title candidate $l_j^{(m)} \in L_m$,
which indicates how relevant it is to the given mention. We represent the mention using various contextual
clues and compute several similarity scores
between the mention and the English title candidates
based on multilingual word and title embeddings. A
ranking model learned from Wikipedia documents is
used to combine these similarity scores and output
the final score for each candidate. We then select
the candidate with the highest score as the answer,
or output NIL if there is no appropriate candidate.

%\subsection{Disambiguation Model}
Formally, we assume that we have access to a snapshot of Wikipedia, in some language $en$\footnote{Deliberately using the symbol $en$ as it is the most widely chosen language in EL research.}, where $en \in X$, $X$ being the set of all languages in Wikipedia,  as our knowledge-base $KB_{en}$ with titles also known as links denoted by ${L_1,\ldots,L_N}$. We can define the goal of Entity Linking (EL) as, given a textual mention $m$ and a document $D$, $m\in D$ and $m,D\in en$, to identify the best link $l_i$: 
\begin{equation}
\hat{l}^{(m)} = \arg\max_{j}P(l_j^{(m)}|m,D) \label{eq1}
\end{equation}
Since computing $P\left(l_j^{(m)}|m,D\right)$ can be prohibitive over large datasets, we change the problem into computing \begin{equation}
\hat{l}^{m}=\arg\max_j P(C|m,D,l_j^{(m)}) \label{hans-eq2}
\end{equation}
where $C$ is a Boolean variable that measures how ``consistent" the pairs $(m,D)$ and $l^{(m)}_j$ are. As a further simplification, given $(m,D)$, we perform an Information Retrieval (IR)-flavored \textit{fast match} to identify the most likely candidate links ${l^{(m)}_{j_1},\ldots,l^{(m)}_{j_m}}$ for the input $(m,D)$, then find the $\arg\max$ over this subset.

In cross-lingual EL, we assume that $m,D \in tr$, where $tr$ is some foreign language like Spanish or Chinese. However, we need to link $m$ to some target link $l^{(m)}_i$, where $l^{(m)}_i \in KB_{en}$.
%The rest of paper is organized as follows: Section \ref{sec:embeddings}
%outlines our the various multilingual
%word and title embeddings that we use in our model for several languages
%in Wikipedia. In Section \ref{sec:repr-details} we present our techniques for modeling contexts using these embeddings. Section \ref{sec:model} presents the proposed neural EL model and how we extend it to perform cross-lingual EL model using the multilingual
%embeddings. Section \ref{sec:experiments} present our experiments. Section \ref{sec:related-work} discusses related
%work and finally, we conclude in Section \ref{sec:conclusion}.

\subsection{Fast Match Search}
\label{sec:fast-match}
The goal of the fast match search is to provide a set of candidates that can be re-scored to compute the $\arg\max$ in Equation (\ref{hans-eq2}). To be able to do this, we prepare an anchor-title index, computed from our Wikipedia snapshot, that
maps each distinct hyper-link anchor text to its target Wikipedia titles \eg\ the anchor text ``Titanic” is used in Wikipedia to refer both to the
famous ship and to the movie. To retrieve the disambiguation candidates $l_i$ for
a query mention $m$, we query the anchor-title index that we constructed. $l_i$
is taken to be the set of titles most frequently linked to with anchor
text $m$ in Wikipedia. For cross-lingual EL, in addition to using the English Wikipedia index (built from the English snapshot), we also build an anchor-title index from the respective target language Wikipedia. Once we have that index, we rely on the inter-language links in Wikipedia to map all the non-English titles back to English. Hence, we have an additional anchor-title index where we have foreign hyper-links as surface forms but English titles as the targets e.g. the surface form ``Estados Unidos'' will have the candidate title \textit{United\_States} which is a title in the English Wikipedia.
\section{Embeddings}
Before delving into the model architecture, we briefly describe the word embeddings used in this work. 
Since we are interested in performing cross-lingual EL, we make use of multi-lingual word embeddings, as shown below.
% Interested readers may read more about them in the cited papers. 
\label{sec:embeddings}
% We first show the technique we use to generate mono-lingual (our case, English) word embeddings. Then we outline three different methods we use to generate multi-lingual embeddings. Finally, we show how we use the mono-lingual word embeddings to generate page embeddings for the english Wikipedia titles.

\subsection{Monolingual Word Embeddings}
\label{sec:mono-emb}
We use the widely used CBOW word2vec model \cite{mikolov2013efficient} to generate English mono-lingual word embeddings.
% We build vector representations of words (word embeddings) for English from monolingual
% data. 
% We adopt a variant of the CBOW word2vec model \cite{mikolov2013efficient},
% which concatenates the context words surrounding
% a target word x using a weight ( $\frac{1}{dist(x;xc)}$ ) that decays
% with the distance of a context word $x_c$ to $x$.
% We train 300-dimensional word embeddings for English. 
% We train word2vec for 1 epoch for English/Spanish and 5 epochs for the rest of the languages for which we have less data.

\subsection{Multi-lingual Embeddings}
\textbf{Canonical Correlation Analysis (CCA):} This technique is based on \cite{faruqui2014improving} who learn vectors by first performing SVD on
text in different languages, then applying CCA on pairs of vectors for the 
words that align in parallel corpora. For cross-lingual EL, we use the embeddings provided by \cite{tsai2016cross}, built using the title mapping obtained from inter-language links in Wikipedia.\\
% Let $A_{en}\in R^{a\times k_1}$ and $A_{tr}\in R^{a\times k_2}$ be the matrices
% containing the embeddings of the aligned English
% and foreign language titles, where $a$ is the number
% of aligned titles and $k_1$ and $k_2$ are the dimensionality
% of English embeddings and foreign language embeddings respectively. They align the embeddings of the two languages using canonical correlation analysis (CCA) as $P_{en}, P_{tr} = CCA(A_{en}, A_{tr})$ where $P_{en}\in R^{k_1\times d}$ and $P_{tr}\in R^{k_2\times d}$ are the projection matrices for English and foreign language embeddings and $d$ is the dimensionality of the projected vectors, which is a parameter in CCA. Let $E_{en}\in R^{n_1\times k_1}$ be the matrix containing monolingual embeddings for all words and titles in English, where the number of words and titles is $n_1$, We obtain the multilingual embeddings of English
% words and titles by $M_{en} =  E_{en}P_{en} \in R^{n_1\times d}$. Similarly, the multilingual embeddings of foreign words and titles are given by $M_{tr} =  E_{tr}P_{tr} \in R^{n_2\times d}$. Here there are $n_2$ words and titles in the foreign language. \\\\
% \begin{equation}
% P_{en}, P_{tr} = CCA(A_{en}, A_{tr})
% \end{equation}
% We use the rows of $M_{tr}$ and $M_{en}$ to build representations of contexts in our model.\\
\textbf{MultiCCA:} Introduced by \cite{ammar2016massively} this technique builds upon CCA and uses a linear operator to project pre-trained monolingual embeddings in each language (except English) to the vector space of pre-trained English word embeddings.
%They let the vector space of the initial (monolingual) English embeddings serve as the multilingual vector space (since English typically offers the largest corpora and wide availability of bilingual dictionaries). Then they estimate projections from the monolingual embeddings of the other languages into the English space.\\
% Recently, \cite{ammar2016massively} extend the idea of \cite{faruqui2014improving} to generate ‘multiCCA’: for pre-training multilingual word embeddings. ‘MultiCCA’ uses a linear operator to project pretrained monolingual embeddings in each language (except English) to the
% vector space of pre-trained English word embeddings. They let the vector space of the initial (monolingual) English embeddings serve as the multilingual vector space (since English typically offers the largest corpora and wide availability of bilingual dictionaries). Then they estimate
% projections from the monolingual embeddings of the other languages into the English space. They assume the projection matrices $P_{en,tr}$ and $P_{tr}$ to be non-singular and embedding dimension to be the same across both languages, $k_1$ = $k_2$ = $d$. They then define the multilingual embedding $M_{en} =  E_{en}\in R^{n_1\times d}$ and $M_{tr} =  E_{tr}P_{tr}P_{en,tr}^{-1} \in R^{n_2\times d}$ for all $tr$ target languages under consideration.\\\\
\\
\textbf{Weighted Least Squares (LS):} Introduced by \cite{mikolov2013exploiting}, the foreign language embeddings are directly projected onto English, with the mapping being constructed through multivariate regression.
\\
\subsection{Wikipedia Link Embeddings} \label{sec:WikiEmbeddings}
We are also interested in embedding entire Wikipedia pages (links). In previous work, \cite{berkeleyNNEL} run CNNs over the entire article and output one fixed-size vector. However, we argue that this operation is too expensive, and it becomes more expensive for some very long pages (based on our experiments on the validation data). We propose a simpler, less expensive route of modeling the Wikipedia page of a target entity.  For every Wikipedia title and using pre-trained word embeddings (obtained in Section \ref{sec:mono-emb}), we compute a weighted average of all the words in the Wikipedia page text. We use the inverse document frequency (IDF) of each word as a weight for its vector, to reduce the influence of frequent words. We compute the Wikipedia page embedding for page $p$ ($e_p$) as:
\begin{equation}
e_p = \dfrac{\sum_{w \in p}{e_{w}idf_w}}{\sum_{w \in p}{idf_w}}
\end{equation}
where $e_w$ and $idf_w$ are the embedding vector and the IDF for word $w$ respectively. 
We further apply (and train) a fully connected \textit{tanh} activation layer to the embedding obtained this way, in order to allow the model to bring the mention context and the Wikipedia link embedding to a similar space before further processing.
\section{Modeling Contexts}
\label{sec:repr-details}
In this Section, we will describe how we build the sub-networks that encode the representation of query mention $m$ in the given query document $D$. This representation is then compared with the page embedding (through cosine similarity) and the result is fed into the higher network (Figure \ref{fig:nn-model}).

Noting that the entire document $D$ might not be useful \footnote{Our experiments on the validation datasets made us think this way.} for disambiguating $m$, we choose to represent the mention $m$ based only on the surrounding sentences of $m$ in $D$, in contrast to \cite{he2013learning,berkeleyNNEL}, which chose to use the entire document for modeling.
% Contrary to modeling the whole source document, which contains the mention, similar to \cite{he2013learning,berkeleyNNEL}, our hypothesis is that the topic of the source document is not only (and always) about the ambiguous mention that we are asked to disambiguate. Instead, we from our training data, we observe that the sentences that contain the mention have the most valuable contextual information.
Hence, following similar ideas in \cite{barrena2014one,lee2012joint}, we run a coreference resolution system \cite{luo2004mention} and assume a ``one link per entity" paradigm (similar to one sense per document \cite{gale1992one,yarowsky1993one}). We then use these to build a \emph{sentence-based} context representation of $m$ as well as its finer-grained context encoding, from only words within a window surrounding the mention occurrences.

\subsection{Modeling Sentences}
\label{sentence-repr}
We collect all the sentences that contain the mention or are part of the entity's coreference chain. Then we combine these sentences together and form a sequence of sentences containing all instances of mention $m$. We use a convolutional neural network (CNN) to produce fixed-size vector representations from the variable length sentences.
We first embed each word into a $d$-dimensional vector space using the embedding techniques described in the previous section . This results in  a sequence of vectors $w_1$,...,$w_n$. We then map those words into a fixed-size vector using a Convolutional Neural Network (CNN) parameterized with a filter bank $V \in \mathbb{R}^{k \times dc}$, where $c$ is the width of the convolution (unigram, bigram, etc.) and $k$ is the number of filter maps. We apply a \textit{tanh} nonlinearity and aggregate the results with \textit{mean}-pooling. A similar CNN is used for building representations of the first paragraphs of a Wikipedia page which is taken to be the context of the candidate link. First paragraphs of an entity's Wikipedia page consists of one or more sentences. Note that this is different than running CNNs on the whole Wikipedia link embeddings described earlier.
% * <raduf@us.ibm.com> 2017-04-11T19:36:31.729Z:
%
% ^.

\subsection{Fine-grained context modeling}
\label{context-repr}
While representing the context of a mention as the output of a CNN running over the sentences surrounding it, might allow for relevant patterns to fire, it is not clear if this type of a representation allows for finer-grained meaning distinctions. Furthermore, this does not exploit the fact that words closer to a mention, are stronger indicators of its meaning than words that are far away. Consider, for example, this sentence: \textit{``Ahmadinejad , whose country has been accused of stoking sectarian violence in Iraq, told ABC television that he did not fear an attack from the United States.''} If our query mention is ABC, only several words surrounding it are needed for a system to infer that ABC is referring to the \textit{American Broadcasting Company} (a television network), while modeling the entire sentence might lead to losing that signal.

For that purpose, we consider \textit{context} to be the words surrounding a mention within a window of length \textit{n}. For our experiments, we chose \textit{n} to be 4. We collect all the left and right contexts separately, the left ending with the mention string and the right beginning with the mention string.
\begin{figure*}
\begin{center}
\includegraphics[width=0.9\textwidth]{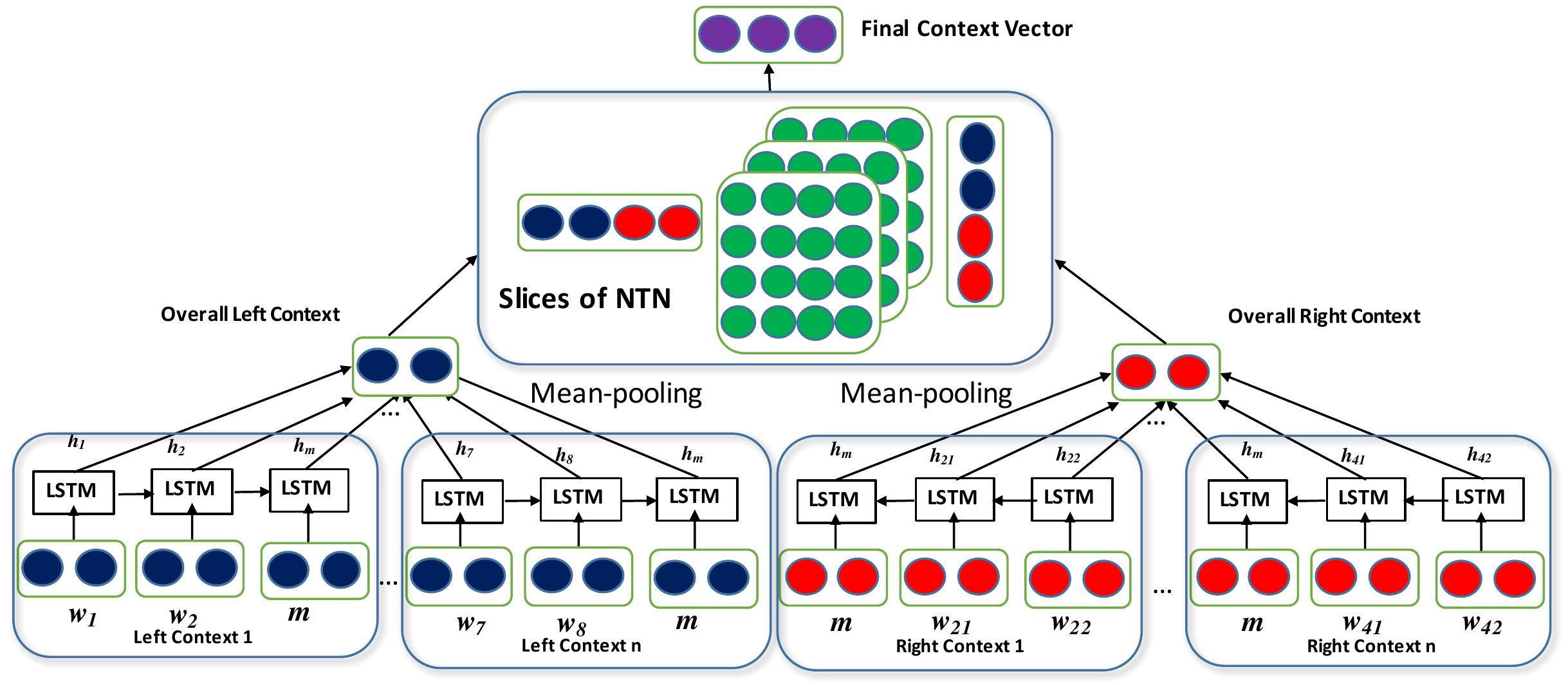}
\end{center}
\caption{Modeling of fine grained context using LSTMs and NTNs from the left and right contexts obtained from the coreference chain of the query entity.}
% \caption{We run forward LSTMs on the left and backward LSTMs on the right contexts (denoted by words $w_i$) of a mention $m$ denoted as a single token for simplicity. The output of the LSTMs are fed a tensor network for semantic compositionality of the contexts and produce a final context representation. The LSTM blocks indicate a standard LSTM memory cell and the NTN (neural tensor network) is shown in Figures \ref{fig:tensor-network1} and \ref{fig:tensor-network2}. \bug need to change the slice fig to 2 by 2.}
\label{fig:lstms}
\end{figure*}

% \begin{figure*}
% \begin{center}
% \includegraphics[trim = 35mm 39mm 55mm 46mm,  width=1.5\columnwidth]{figs/lstm}
% \end{center}
% \caption{We run forward LSTMs on the left and backward LSTMs on the right contexts (denoted by words $w_i$) of a mention $m$. Of course, $m$ can have multiple tokens but for simplicity we denote it as a single token. We put the output of the LSTMs onto a tensor network for semantic compositionality of the contexts and produce a final context representation. The LSTM blocks indicate a standard LSTM memory cell and the NTN (neural tensor network) is shown in Figures \ref{fig:tensor-network1} and \ref{fig:tensor-network2}.}
% \label{fig:lstms}
% \end{figure*}

% \begin{figure}[t]
% \begin{center}
% % \fbox{\includegraphics[trim = 75mm 63mm 92mm 45mm, clip,width=0.9\columnwidth]{figs/tensor}}
% \fbox{\includegraphics[trim = 5mm 30mm 10mm 30mm, clip,width=\columnwidth]{figs/tensor2}}
% \end{center}
% \caption{We show the visualization of our Neural Tensor Network (inspired by \protect\cite{socher2013reasoning}) . Each dashed box represents one slice of the tensor, in this case there are $k = 2$ slices. }
% \label{fig:tensor-network1}
% \end{figure}

% \begin{figure}[t]
% \begin{center}
% % \fbox{\includegraphics[trim = 75mm 63mm 92mm 45mm, clip,width=0.9\columnwidth]{figs/tensor}}
% \fbox{\includegraphics[scale = 0.5,trim = 75mm 63mm 92mm 45mm, clip]{figs/tensor}}
% \end{center}
% \caption{An illustration of a neural tensor network for modeling the semantic composition between a  mention's left and right contexts.}
% \label{fig:tensor-network2}
% \end{figure}
In a first step, we run LSTMs on these contexts as follows: we run forward LSTMs on the left and backward on the right context and use element-wise mean pooling as the combination strategy. To detail: 
using the condensed notations of \cite{DBLP:journals/corr/ChengDL16}, we run a forward LSTM network over each left context, and a backward LSTM network over each right context, and pool them over all the contexts of each mention. The resulting condensed representations are averaged and then combined using a neural tensor network, using the equation below (also see Figure \ref{fig:lstms}).
\begin{equation}
NTN(l,r;W) = f(\left[ \begin{array}{c}l \\ r \\\end{array}\right]^t W^{\{1,\ldots,k\}}\left[ \begin{array}{c}l \\ r \\\end{array}\right])
\label{eq:NTN}
\end{equation}
Here $l$ and $r$ are the representations for the overall left and right context ($l,r \in \mathbb{R}^d$), $W$ is a tensor with $k$ slices with $W^i \in \mathbb{R}^{2d \times 2d}$, $f$ is a standard nonlinearity applied element wise (sigmoid in our case). The output of NTN is a vector $NTN(l, r;W) \in \mathbb{R}^k$ \footnote{We use $l$ to denote left context here for simplicity even when we have used it before to denote a link.}.

\section{Cross-Lingual Neural Entity Linking}
\label{sec:model}

\subsection{Neural Model Architecture}
\begin{figure}[t]
\begin{center}
\fbox{\includegraphics[trim = 1mm 2mm 0mm 1mm, clip,width=\columnwidth]{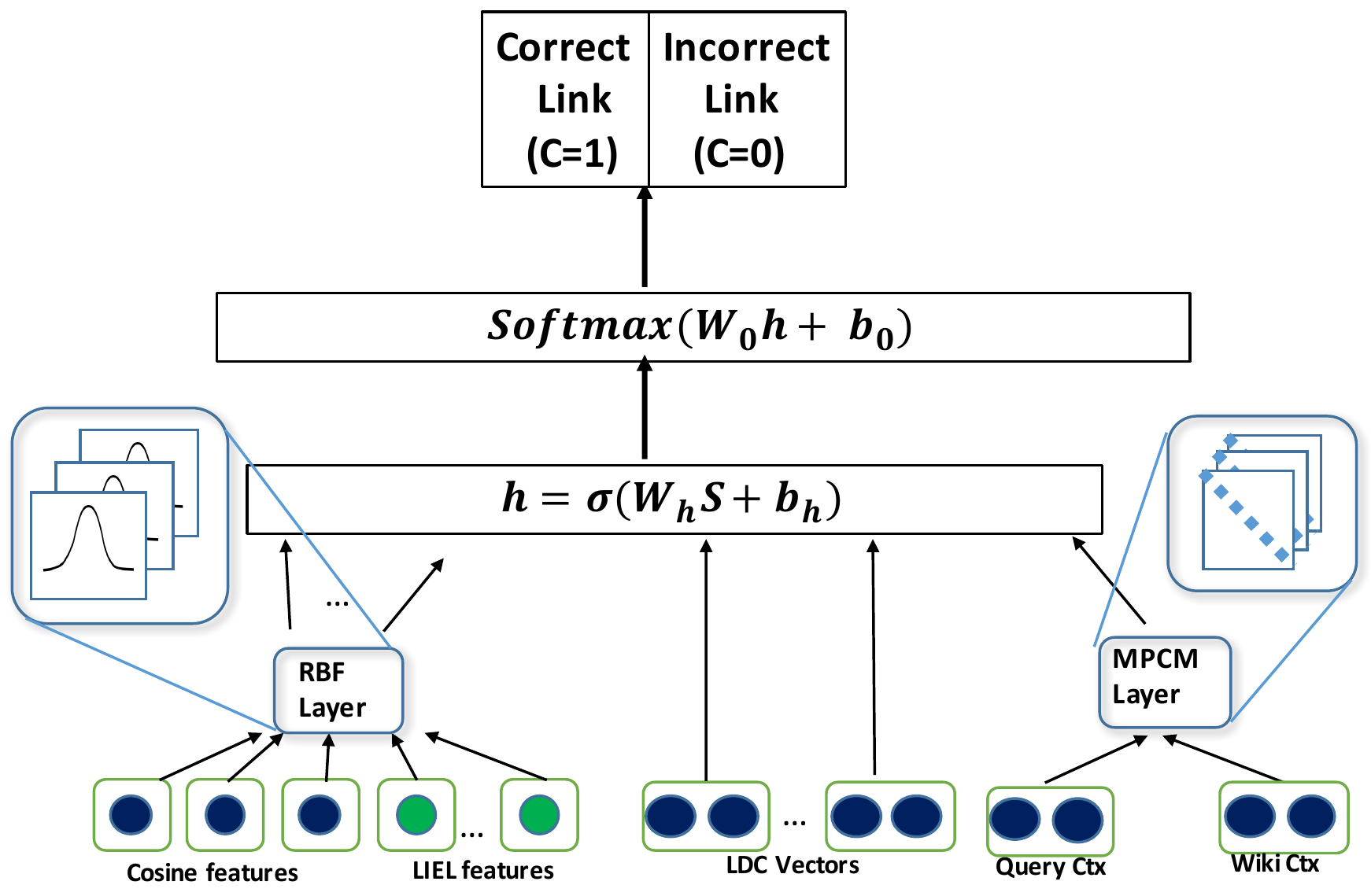}}
% \fbox{\includegraphics[trim = 40mm 20mm 60mm 50mm, clip,width=\columnwidth]{figs/model_new}}
\end{center}
\caption{\textbf{Architecture of our neural EL system.} The input to the system are: a document $D$ containing the query mention $m$ and the corresponding Wikipedia candidate link $l_i\in L$, where $L$ is the set of all possible links extracted from the fast match step described in Section \ref{sec:fast-match}.}
\label{fig:nn-model}
\end{figure}

The general architecture of our neural EL model is described in Figure \ref{fig:nn-model}.  Our target is to perform ``zero shot learning'' \cite{socher2013zero,palatucci2009zero} for cross-lingual EL. Hence, we want to train a model on English data and use it to decode in any other language, provided we have access to multi-lingual embeddings from English and the target language. We allow the model to compute several similarity/coherence \textit{scores} $S$ (feature abstraction layer): which are several measures of similarity of the context of the mention $m$ in the query document and the context of the candidate link's Wikipedia page, described in details in the next section, which are fed to a feed-forward neural layer $H$ with weights $W_h$, bias $b_h$, and a \textit{sigmoid} non-linearity. The output of $H$ (denoted as $h$) is computed according to $h = \sigma(W_h S + b_h)$.
The output of the binary classifier $p(C|m,D,l)$ is the softmax over the output of the final feed-forward layer $O$ with weights $W_0$ and bias $b_0$. $p(C|m,D,L)$ represents the probability of the output class C taking a value of 1 (correct link) or 0 (incorrect link), and is computed as a 2 dimensional vector and given by:
% \vspace{-2mm}
\begin{equation}
p(C|m,D,l) = softmax(W_0h+b_0)
\end{equation}
%\vspace{-2mm}
\subsection{Feature Abstraction Layer}
\label{sec:feature-layer}
In this layer, we encode the similarity between the context of the mention in the source document and the context of the corresponding candidate Wikipedia links as obtained through fast match at multiple granularities, described below. \\\\
% Previous neural EL research \cite{sun2015modeling,he2013learning} use only one similarity feature as a cost function in their models. In contrast, our hypothesis is that one can perform better disambiguation by comparing contextual information at different granularity from the source document and the Wikipedia candidate's page.
% Additionally, given a mention, we extend its \emph{relevant context} to consist not only of sentences surrounding it but also sentences surrounding other (within-document) mentions of the same entity, according to the output of a coreference resolution system. (details in Section \ref{repr-details}). 
% We briefly outline our similarity-based features below and describe them at length in Section \ref{similarityscores}.\\
\textbf{A. Similarity Features by comparing Context Representations}\\
%\begin{itemize}
%\item
\textbf{1. ``Sentence context - Wiki Link" Similarity}:
The first input to this layer is the cosine similarity between the CNN representations of its \emph{relevant context} sentences and the embedding of the candidate Wikipedia link (both described in the Embeddings section).\\ 
%\item
\textbf{2. ``Sentence context - Wiki First Paragraph'' Similarity}: The next input is the cosine similarity between the CNN representations of the sentential context of a mention and the first Wikipedia paragraph, following the intuition that often the first paragraph is a concise description of the main content of a page. Multiple sentences are composed using the same model as above.\\
%\item
\textbf{3. ``Fine-grained context - Wiki Link'' Similarity}:
Next, we feed the similarity between the more fine-grained embedding of context described in the Embeddings section, Equation (\ref{eq:NTN}) and the embedding of the candidate page link.\\
%\end{itemize}
\textbf{4. Within-language Features}: We also feed in all the local features described in the LIEL system \cite{sil2016one}. LIEL uses several features such as ``how many words overlap between the mention and Wikipedia title match?'' or ``how many outlink names of the candidate Wikipedia title appear in the query document?'' that compares the similarity of the context of the entity under consideration from the source document and its target Wikipedia page. We also add a feature encoding the probability $P(l_i|m)$, the posterior of a Wikipedia title $l_i$ being the target page for the mention $m$, using solely the anchor-title index. This feature is a strong indicator to predict if a link $l_i$ is the correct target for mention $m$.\\
\textbf{Multi-perspective Binning Layer:} Previous work \cite{liu2016neural} quantizes numeric feature values and then embeds the resulting bins into 10-dimensional vectors. In contrast, we propose a ``Multi-perspective Binning Layer'' (MPBL) which applies multiple Gaussian radial basis functions to its input, which can be interpreted as a smooth binning process. The above-described similarity values are fed into this MPBL layer, which maps each to a higher dimensional vector. Introducing this layer lets the model learn to respond differently to different values of the cosine input feature, in a neural network friendly way. Our technique differs from \cite{liu2016neural} in that it is able to automatically learn the important regions for the input numeric values.\\[12pt]
\textbf{B. Semantic Similarities and Dissimilarities}\\
%\begin{itemize}
%\item
\textbf{1. Lexical Decomposition and Composition (LDC)}: We use the recently proposed LDC\footnote{Not to be confused with the Linguistic Data Consortium (https://www.ldc.upenn.edu/.)} model in \cite{wang2016sentence} to compare the contexts. For brevity, we only give a brief description of this feature - we direct the reader to the original paper. We represent the source context \ct{} and the Wikipedia paragraph \wpf{} as a sequence of pre-trained embeddings of words. \ct{} = [$\cti{}_1$, \ldots, $\cti{}_m$] and \wpf{}=[$\wpfi{}_1$, \ldots, $\wpfi{}_n$] where $\cti{}_i$ and $\wpfi{}_j$ are the pre-trained word embeddings for the $i$th and $j$th word from the source context and the Wikipedia paragraph respectively.
The steps of LDC are summarized below.
For each word $\cti{}_i$ in $\ct{}$, the semantic matching step finds a matching word $\hat{\cti{}_i}$ from $\wpf{}$. In the reverse direction, a matching word $\hat{\wpfi{}_j}$ is found for each $\wpfi{}_j$ in $\wpf{}$. For a word embedding, its matching word is the one with the highest cosine similarity. Hence, $\hat{\cti{}_i} = \wpfi{}_k \mbox{ where k = }\argmax_j \cosine(\cti{}_i,\wpfi{}_j)$ and $\hat{\wpfi{}_j} = \cti{}_k \mbox{ where k = }\argmax_i \cosine(\wpfi{}_j, \cti{}_i)$.
% \vspace{-2mm}
\begin{equation*}
\hat{\cti{}_i} = \wpfi{}_k \mbox{ where k = }\argmax_j \cosine(\cti{}_i,\wpfi{}_j) 
\end{equation*}
\begin{equation*}
\hat{\wpfi{}_j} = \cti{}_k \mbox{ where k = }\argmax_i \cosine(\wpfi{}_j, \cti{}_i)
\end{equation*}
The next step is decomposition, where each word embedding $\cti{}_i$ (or $\wpfi{}_j$) is decomposed based on its semantic matching vector $\hat{\cti{}_i}$ (or $\hat{\wpfi{}_j}$) into two components: similar component $\cti{}_i^{+}$ (or $\wpfi{}_j^{+}$) and dissimilar component $\cti{}_i^{-}$(or $\wpfi{}_j^{-}$). We compute the cosine similarity between $\cti{}_i$ and $\hat{\cti{}_i}$ (or $\wpfi{}_i$ and $\hat{\wpfi{}_i}$) and decompose linearly. Hence, $(\cti{}_i^{+},\cti{}_i^{-})=(\alpha \cti{}_i, (\sqrt{1-\alpha^2}) \cti{}_i)$ and $(\wpfi{}_i^{+},\wpfi{}_i^{-})=(\alpha \wpfi{}_i, \sqrt{1-\alpha^2} \wpfi{}_i)$ where $\alpha = \cosine(\cti{}_i,\hat{\cti{}_i})$ and $\alpha = \cosine(\wpfi{}_i,\hat{\wpfi{}_i})$.
% \vspace{-2mm}
\begin{equation*}
(\cti{}_i^{+},\cti{}_i^{-})=(\alpha \cti{}_i, (1-\alpha) \cti{}_i) \mbox{ and } \alpha = \cosine(\cti{}_i,\hat{\cti{}_i})
\end{equation*}
\begin{equation*}
(\wpfi{}_i^{+},\wpfi{}_i^{-})=(\alpha \wpfi{}_i, (1-\alpha) \wpfi{}_i) \mbox{ and } \alpha = \cosine(\wpfi{}_i,\hat{\wpfi{}_i})
\end{equation*}
In the Composition step, the similar and dissimilar components are composed at different granularities using a two channel CNN and pooled using \textit{max}-pooling. The output vector is the representation of the similarity (and dis-similarity) of the source context of the mention with the Wikipedia page of the target entity.\\
\textbf{2. Multi-perspective Context Matching (MPCM)}: Next, we input a series of weighted cosine similarities between the query mention context and the Wikipedia link embedding, as described in \cite{wang2016multi}. Our argument is that while cosine similarity finds semantically similar words, it has no relation to the classification task at hand. Hence, we propose to train weight vectors to re-weigh the dimensions of the input vectors and then compute the cosine similarity. The weight vectors will be trained to maximize the performance on the entity linking task.
We run CNNs to produce a fixed size representations for both query and candidate contexts from Section \ref{sec:repr-details}. We build a node computing the cosine similarity of these two vectors, parametrized by a weight matrix. Each row in the weight matrix is used to compute a score as $u_k = \cosine(w_k \circ v_1, w_k \circ v_2)$, where $v_1$ and $v_2$ are input $d$ dimensional vectors, $w_k$ $\in$ $\mathbb{R}^{d}$ is the $k^{\textnormal{th}}$ column in the matrix, $\mathbf{u}$ is a $l$-dimensional output vector, and $\circ$ denotes a element-wise multiplication. Note that re-weighting the input vectors is equivalent to applying a diagonal tensor with non-negative diagonal entries to the input vectors.\\
% hans' edit for brevity
% \textbf{Multi-perspective Context Matching}: Next, we input a series of weighted cosine similarities between the query mention context and the Wikipedia link embedding, as described in \cite{wang2016multi}. 
% % While cosine similarity finds semantically similar words, it has no relation to the classification task at hand. Hence, we propose to train weight vectors to re-weigh the dimensions of the input vectors and then compute the cosine similarity. The weight vectors will be trained to maximize the performance on the entity linking task.
% We run CNNs to produce a fixed size representations for both query and candidate contexts from Section \ref{sec:repr-details}. We build a node computing the cosine similarity of these two vectors, parametrized by a weight matrix. Each row in the weight matrix is used to compute a score as $u_k = \cosine(w_k \circ v_1, w_k \circ v_2)$, where $v_1$ and $v_2$ are input $d$ dimensional vectors, $w_k$ $\in$ $\mathbb{R}^{d}$ is the $k^{\textnormal{th}}$ column in the matrix, $\mathbf{u}$ is the $l$-dimensional output vector, and $\circ$ denotes the element-wise multiplication. Note that re-weighting the input vectors is equivalent to applying a diagonal tensor with non-negative diagonal entries to the input vectors.\avi{example here as well. Bring back formula}\\
% end hans' edit for brevity
%
%
% Since these features are typically scalar values we feed them individually into the RBF layer as above and map them into 100-D vector for each feature.
%
\subsection{Training and Decoding}
To train the model described in Equation (\ref{hans-eq2}), the binary classification training set is prepared as follows. For each mention $m_{ij}\in D_i$ and its corresponding correct Wikipedia page $l^{(m_{ij})}$, we use our fast match strategy (discussed in Page 2) to generate $K_{ij}$ number of incorrect Wikipedia pages $(l_{ij_k})_k$. $l_i$ and $l_{ij_k}$ represent positive and negative examples for the binary classifier. Pairs in the list of $[(m_{ij},D,l_{ij}),(m_{ij},D,l_{ij_0}),\ldots,(m_{ij},D_i,l_{ij_{K_{ij}}})]$ will be used to produce the similarity/ dis-similarity vectors $S_{ij_k}$. Classification label $Y_{ij_k}$ that corresponds to input vector $(m_{ij},D_{i},l_{ij_k})$ will take the value of $1$ for the correct Wikipedia page and $0$ for incorrect ones. The binary classifier is trained with the training set $T$ which contains all the $(m,D,l,Y)$ data pairs\footnote{The ratio of positive to negative training events is controlled  to produce a more balanced training data.}. 

Training is performed using stochastic gradient descent on the following loss function:
\begin{equation}
-\frac{1}{|T|}\sum_{(m_j,D_j,l_j,Y_j)\in T}\log P(C=Y_j|m_j,D_j,l_j)
\end{equation}

Decoding a particular mention $m\in D$, is simply done by running fast match  to produce a set of likely candidate Wikipedia pages, then generate the system output $\hat{l^{(m)}}$ as in Equation (\ref{hans-eq2}).\\
Note that the model does all this by only computing similarities between texts in the same language, or by using cross-lingual embeddings, allowing it to transcend across languages.
\section{Experiments}
\label{sec:experiments}
We evaluate our proposed method on the benchmark datasets for English: CoNLL 2003 and TAC 2010 and Cross-Lingual: TAC 2015 Trilingual Entity Linking dataset.

% For the English experiments, we use the CBOW word2vec model as shown in \cite{mikolov2013distributed}. We use $\approx$ 8 billion tokens as input (a concatenation of Wikipedia, Gigaword and the Bolt corpora) and the following word2vec settings: negative sampling (10 samples), context window of 5, subsampling set to 1e05 and 1 iteration.

% For the multi-lingual experiments, we experiment with two methods—
% ‘multiCCA’ and ‘multiCluster,’ both proposed by \cite{ammar2016massively} which extends the idea of \cite{faruqui2014improving} much further: for pre-training multilingual word embeddings. ‘MultiCCA’ uses a linear
% operator to project pretrained monolingual embeddings
% in each language (except English) to the
% vector space of pre-trained English word embeddings,
% while ‘multiCluster’ uses the same embedding
% for translationally-equivalent words in different languages. On our experiments, we saw multiCCA to be performing better on the development data and we hence report our final performance using that setting. \bug We also perform experiments using the UIUC cross-lingual embeddings \cite{tsai2016cross} which follow \cite{faruqui2014improving}.

\subsection{Datasets}
\textbf{English} (CoNLL \&\ TAC): The CoNLL dataset \cite{2011-emnlp-NE-disambig-yago} contains 1393 articles with about 34K mentions,
and the standard performance metric is
mention-averaged accuracy. The documents are
partitioned into train, test-a and test-b. Following previous work, we report performance on the 231 test-b
documents with 4483 linkable mentions. The TAC 2010 source collection includes news
from various agencies and web log data. Training
data includes a specially prepared set of 1,500
web queries. Test data includes 2,250 queries –
1,500 news and 750 web log uniformly distributed
across person, organisation, and geo-political entities. \\
\textbf{Cross-Lingual}   (TAC): We evaluate our method on the TAC 2015 Tri-Lingual Entity Linking datasets which comprises of 166 Chinese documents (84 news and 82 discussion forum articles) and 167 Spanish documents (84 news
and 83 discussion forum articles). The mentions in
this dataset are all named entities of five types: Person,
Geo-political Entity, Organization, Location,
and Facility.

% We use standard train, dev and test splits if the datasets come with it. We also use the Wikipedia dataset \cite{2011-acl-illinois-wikifier} as part of the training (and dev) data as we it is known that deep Neural Nets perform better with more data.
We use standard train, validation and test splits if the datasets come with it, else we use the CoNLL validation data as dev. For the CoNLL experiments, in addition to the Wikipedia anchor-title index, we also use a alias-entity mapping previously used by \cite{pershina2015personalized,globerson2016collective,yamada2016joint}. We also use the mappings provided by \cite{2011-emnlp-NE-disambig-yago} obtained by extending the “means” tables
of YAGO \cite{hoffart2013yago2}.
% For the CoNLL experiments, we also use a alias-entity mapping provided by \cite{pershina2015personalized} and previously used by \cite{globerson2016collective,yamada2016joint}. We also use the mappings provided by \cite{2011-emnlp-NE-disambig-yago} obtained by extending the “means” tables
% of YAGO \cite{hoffart2013yago2}.

\subsection{Hyperparameters}
We tune all our hyper-parameters on the development data. We run CNNs on the sentences and the Wikipedia embeddings with filter size of 300 and width 2. The non-linearity used is \textit{tanh}. For both forward (left) and backward (right) LSTMs, we use mean pooling. We tried max-pooling and also choosing the last hidden state of the LSTMs but mean pooling worked the best. We combine the LSTM vectors for all the left and all the right using mean pooling, as well. For the NTNs, we use sigmoid as the non-linearity and an output size of 10 and use $L2$ regularization with a value of 0.01. Finally, to compute the similarity we feed the output of the NTN to another hidden layer with sigmoid non-linearity for a final output vector of size 300. For the main model, we again use sigmoid non-linearity and an output size of 1000 with a dropout rate of 0.4. We do not update the Wikipedia page embeddings as they did not seem to provide gains in numbers while testing on development data. We also do not update the multi-lingual embeddings for the cross-lingual experiments. For the English experiments, we update the mono-lingual English word embeddings. For the MPBL node, the number of dimensions is 100.

\subsection{Comparison with the SOTA}
The current SOTA for English EL are \cite{globerson2016collective} and \cite{yamada2016joint}. We also compare with LIEL \cite{sil2016one} which is a language-independent EL system and has been a top performer in the TAC annual evaluations. For cross-lingual EL, our major competitor is \cite{tsai2016cross} who uses multi-lingual embeddings similar to us. We also compare with several other systems as shown in Table \ref{table:conll}, \ref{table:tac-en} and \ref{table:tac-es} along with the respective top ranked TAC systems.

\begin{table*}[t]
\centering
\small
\subfloat[t][CoNLL2003] {
\begin{tabular}{lcc}
\toprule
Systems	& In-KB acc. \%		\\
\midrule
Hoffart \etal\ (2011)   &82.5\\
Gupta \etal (2017) & 82.9\\
He \etal\ (2013)   &85.6\\
Francis-Landau \etal\ (2016)   &85.5\\
Sil \& Florian (2016)   &86.2\\
Lazic \etal\ (2015)   &86.4\\
Chisholm \& Hachey (2015)   &88.7\\
Ganea \etal\ (2015)   &87.6\\
Pershina \etal\ (2015) &91.8\\
Globerson \etal\  (2016)   &92.7\\
Yamada \etal\ (2016)  &93.1\\
\midrule
\small
This work  &92.1\\
This work+CtxLSTMs  &93.0\\
This work+CtxLSTMs+LDC   &93.4\\
This work+CtxLSTMs+LDC+MPCM   &\textbf{94.0}\\
\bottomrule
\end{tabular}
%\caption{CoNLL2003}
\label{table:conll}
}
\subfloat[t][TAC2010]{
\begin{tabular}{lcc}
\toprule
Systems	& In-KB acc. \%		\\
\midrule
TAC Rank 1   &79.2\\
TAC Rank 2   &71.6\\
Sil \& Florian (2016)   &78.6\\
He \etal\ (2013)  &81.0\\
Chisholm \& Hachey (2015)   &80.7\\
Sun \etal\ (2015)	&83.9\\
Yamada \etal\ (2016)	&85.2\\
Globerson \etal\  (2016)	&87.2\\
\midrule
\small
This work   &85.0\\
This work+CtxLSTMs   &86.3\\
This work+CtxLSTMs+LDC   &86.9\\
This work+CtxLSTMs+LDC+MPCM   &\textbf{87.4}\\
\bottomrule
\end{tabular}
\label{table:tac-en}
}
\caption{\textbf{Performance comparison on the CoNLL 2003 testb and TAC2010 datasets.} Our system outperforms all EL systems, including the only other multi-lingual system, \cite{sil2016one}.}
\end{table*}

\subsection{English Results}
Table \ref{table:conll} shows our performance on the CoNLL dataset along with recent competitive systems in terms of micro-average accuracy. We outperform \cite{globerson2016collective} by an absolute average of 1.27\% and \cite{yamada2016joint} by 0.87\%. Globerson \etal\ use a multi-focal attention model to select specific context words that are essential for linking a mention. Our model  with the lexical decomposition and composition and the multi-perspective context matching layers seems to be more beneficial for the task of EL.
%Interestingly, we also outperform \cite{berkeleyNNEL} which runs CNNs over multiple granularities similar to this work, which suggests that having features computed by running LSTMs are better than CNNs alone.\\

%To test the robustness of our model on a diverse genre of data, we also compare it with some of the other state-of-the-art systems on the benchmark TAC 2010 dataset. 

Table \ref{table:tac-en} shows our results when compared with the top systems in the evaluation along with other SOTA systems on the TAC2010 dataset. Encouragingly, our model’s performance is slightly better than the top performer, Globerson (2016), and outperforms both the top rankers from this challenging annual evaluation by 8\% absolute percentage points. Note that in both the datasets, our model obtains 7.77\% (on CoNLL) and 8.75\% (on TAC) points better than \cite{sil2016one}, which is a SOTA multi-lingual system. Another interesting fact we observe is that our full model outperforms  \cite{sun2015modeling} by 3.5\% points, where they employ NTNs to model the semantic interactions between the context and the mention. Our model uses NTNs to model the left and right contexts from the full entity coreference chain in a novel fashion not used previously in the EL research and seems highly useful for the task.
%and is a top ranking system in the recent annual TAC evaluations. 
Interestingly, we observe that the recent \cite{gupta2017entity} EL system performs rather poorly on the CoNLL dataset (7.5\% lower than our model) even when their system employ entity type information from a KB which our system does not.

While doing ablation study, we notice that adding the LDC layer provides a boost to our model in both the datasets, and the multi-perspective context matching (MPCM) layer provides an additional  0.5\% (average) points improvement. We see that adding in the context LSTM based layer (fine-grained context) adds almost 1\% point (in both the datasets) over the base similarity features.

\subsection{Cross-lingual Results}
\textbf{Spanish:} Table \ref{table:tac-es} shows our performance on cross-lingual EL on the TAC2015 Spanish dataset. The experimental setup is similar as in the TAC diagnostic evaluation, where systems need to predict a link as well as produce the type for a query mention. We use an entity type classifier to attach the entity types to the predicted links as described in our previous work in \cite{sil2015ibm}. We compare our performance to \cite{sil2016one}, which was the top ranked system in TAC 2015, and the cross-lingual wikifier \cite{tsai2016cross}. We see that our zero-shot model trained with the multi-CCA embeddings is 1.32\% and 1.85\% percentage points better than the two competitors respectively.\\
% Interestingly, our model with the multi-lingual embeddings obtained by the least squares technique is better than the competitors but is lower than the Multi-CCA embeddings by 0.4\% points.\\
\textbf{Chinese:} Table \ref{table:tac-zh} displays our performance on the TAC2015 Chinese dataset. 
%Tsai and Roth is slightly better than the top ranker in the evaluation. 
Our proposed model is 0.73\% points better than \cite{tsai2016cross}. In both cross-lingual experiments, the multi-CCA embeddings outperform LS and CCA methods. In Spanish, LS and CCA are tied but in Chinese, CCA performs better than LS. Note that ``this work'' in Table 2 indicates our full model with LDC and MPCM.
\begin{table*} 
%\begin{center}
\centering
\small
\subfloat[t][Spanish]{
%\resizebox{\columnwidth}{!}{
\begin{tabular}{lcc}
\toprule
Systems	& Linking Acc \%		\\
\midrule
Sil \& Florian (2016) / TAC Rank 1   &80.4\\
Tsai \& Roth (2016)  &80.9\\
\midrule
This Work (LS)   &81.9\\
This Work (CCA)   &81.8\\
This Work (Multi-CCA)   &\textbf{82.3}\\
\bottomrule
\end{tabular}
%}
%\end{center}
\label{table:tac-es}
}\quad \quad
%\end{table}
%\begin{table} 
%\begin{center}
%\resizebox{\columnwidth}{!}{
\subfloat[t][Chinese] {
\small
\begin{tabular}{lcc}
\toprule
Systems	& Linking Acc \%		\\
\midrule
TAC Rank 1   &83.1\\
Tsai \& Roth (2016)  &83.6\\
\midrule
This Work (LS)   &84.1\\
This Work (CCA)   &84.3\\
This Work (Multi-CCA)   &\textbf{84.4}\\
\bottomrule
\end{tabular}
\label{table:tac-zh}
%}
%\end{center}
}
\caption{\textbf{Performance comparison on the TAC 2015 Spanish and Chinese datasets.} Our system outperforms all the previous EL systems.}
%\caption{\textbf{Performance comparison on the TAC 2015 Chinese dataset.}}

\end{table*}
\section{Related Work\label{sec:related-work} }

% Entity linking (EL) has been a central task for the
% NIST-organized Text Analysis Conference since 2009, in the Knowledge
% Base Population track \cite{ji2014overview}. 
Previous works in EL \cite{2006-eacl-bunescu-wikipedia-ne-disambig,2007-cikm-mihalcea-wikify} involved finding the similarity of the context in the source document and the context of the candidate Wikipedia titles. Recent research
on EL has focused on sophisticated global disambiguation algorithms \cite{globerson2016collective,2008-cikm-milne-witten-wikifier,emnlp-chengroth-2013,cikm-joint-nerel} but are more expensive since they capture coherence among titles in the given document. 
However, \cite{2011-acl-illinois-wikifier} argue that global systems provide a minor improvement over local systems. Our proposed EL system is a local system which comprises of a deep neural network architecture with various layers computing the semantic similarity of the source documents and the potential entity link candidates modeled using techniques like neural tensor network, multi-perspective cosine similarity and lexical composition and decomposition.

Sun \etal\ (2015) used neural tensor networks for entity linking, between mention and the surrounding context. But this did not give good results in our case. Instead, the best results were obtained by composing the left and right contexts of all the mentions in the coreference chain of the target mention. In this work, we also introduced state-of-the-art similarity models like MPCM and LDC for entity linking. Combination of all these components helps our model score 3.5 absolute accuracy improvement over Sun \etal\ (2015).

The cross-lingual evaluation at TAC KBP EL Track that started in 2011 \cite{jioverview2011,ji2015overview} has Spanish and Chinese as the 
target foreign languages. One of the top performers \cite{sil2016one}, like most other participants, perform EL in the foreign language (with the corresponding foreign KB), and then find the corresponding English titles using Wikipedia inter-language links. Others \cite{2011cross} translate the query documents to English and do English EL. The first approach relies on a large enough KB in the foreign language, whereas the second depends on a good machine translation system. Similar to \cite{tsai2016cross}, the ideas proposed in this paper make significantly simpler assumptions on the availability of such resources, and therefore can also scale to lower resource languages, while doing very well also on high-resource languages. However, unlike our model they need to train and decode the model on the target language. Our model once trained on English can perform cross-lingual EL on any target language.

Some recent work involves \cite{lin2017list} but is unrelated since it solves a different problem (EL from only lists) than generic EL and hence an apples-apples comparison cannot be done. \cite{pan2017cross} is related but their method prefers common popular entities in Wikipedia and they select training data based on the topic of the test set. Our proposed method is more generic and robust as it is once trained on the English Wikipedia and tested on any other language without re-training. \cite{tan2017entity} solves a different problem by performing EL for queries while we perform EL for generic documents like news. Recently \cite{gupta2017entity} propose an EL system by jointly encoding types from a knowledge-base. However, their technique is limited to only English and unlike us do not perform cross-lingual EL.

\section{Conclusion}

\label{sec:conclusion}
Recent EL research, that we compare against, have produced models that achieve either SOTA mono-lingual performance or cross-lingual performance, but not both. We produce a model that performs zero-shot learning for the task of cross-lingual EL: once trained on English, the model can be applied to any language, as long as we have multi-lingual embeddings for the target language. Our model makes effective use of %deep NNs - mixing of CNNs and LSTMs producing context representations that are effective at capturing the
the similarity models (LDC, MPCM) and composition methods (neural tensor network) to capture similarity/dissimilarity between the query mention's context and the target Wikipedia link's context. We test three methods of generating multi-lingual word embeddings and determine that the MultiCCA-generated embeddings perform best for the task of EL for both Spanish and Chinese. Our model has strong experimental results, outperforming all the previous SOTA systems in both mono and cross-lingual experiments. Also, with the increased focus on cross-lingual EL in future TAC evaluations, we believe that this zero-shot learning technique would prove useful for low-resource languages: train one model and use it for any other language.

\section{Acknowledgements}
We thank Zhiguo Wang for the help with the LDC and MPCM node. We also thank Georgiana Dinu and Waleed Ammar for providing us with the multi-lingual embeddings. We are grateful to Salim Roukos for the helpful discussions, and the anonymous reviewers for their suggestions.
% The first author would like to dedicate his contribution to Donut (his cat) who woke him up in the middle of the night to continue the experiments.

% Add additional packages here. The following
% packages may NOT be used  (this list
% is not exhaustive:
% authblk, caption, CJK, float, fullpage, geometry, 
%hyperref, layout, nameref, natbib, savetrees, 
%setspace, titlesec, tocbibind, ulem
%
%US Lettersize Paper Is Required

%
%
% PDFINFO
% You are required to complete the following
% for pass-through to the PDF. 
% No LaTeX commands of any kind may be
% entered. The parentheses and spaces 
% are an integral part of the 
% pdfinfo script and must not be removed.
%
\bibliographystyle{aaai}
\bibliography{sil}

\end{document}